# Acoustic SLAM based on the Direction-of-Arrival and the Direct-to-Reverberant Energy Ratio

Wenhao Qiu, Gang Wang, and Wenjing Zhang

*Abstract*—This paper proposes a new method that fuses acoustic measurements in the reverberation field and low-accuracy inertial measurement unit (IMU) motion reports for simultaneous localization and mapping (SLAM). Different from existing studies that only use acoustic data for direction-of-arrival (DoA) estimates, the source's distance from sensors is calculated with the direct-to-reverberant energy ratio (DRR) and applied as a new constraint to eliminate the nonlinear noise from motion reports. A particle filter is applied to estimate the critical distance, which is key for associating the source's distance with the DRR. A keyframe method is used to eliminate the deviation of the source position estimation toward the robot. The proposed DoA-DRR acoustic SLAM (D-D SLAM) is designed for three-dimensional motion and is suitable for most robots. The method is the first acoustic SLAM algorithm that has been validated on a real-world indoor scene dataset that contains only acoustic data and IMU measurements. Compared with previous methods, D-D SLAM has acceptable performance in locating the robot and building a source map from a real-world indoor dataset. The average location accuracy is 0.48 m, while the source position error converges to less than 0.25 m within 2.8 s. These results prove the effectiveness of D-D SLAM in real-world indoor scenes, which may be especially useful in search and rescue missions after disasters where the environment is foggy, i.e., unsuitable for light or laser irradiation.

*Index Terms*—Simultaneous localization and mapping, robot audition, direct-to-reverberant energy ratio, mobile robots

## I. INTRODUCTION

RECENTLY, there has been renewed interest in simultaneous localization and mapping (SLAM). Many meaningful and excellent works in SLAM have been based on optical and visual sensors, such as VINS [1]. Different from visual SLAM, some works have focused on acoustic SLAM where acoustic sensors are involved. Most works on acoustic SLAM have been conducted in underwater environments [2], [3], [4], while indoor acoustic SLAM [5], [6], [7], by contrast, has received scant attention.

Based on the sensor type used, indoor acoustic SLAM can be classified as active or passive acoustic SLAM. Active indoor acoustic SLAM is usually based on active sonar, and a sonar beam is utilized in an active sonar sensor model to measure the positions of landmarks. To [1]assist in localization, a motion sensor is required to generate motion reports. Passive indoor acoustic SLAM is usually based on microphone arrays for direction-of-arrival (DoA) estimates and motion sensors (such as an odometer) for motion reports. In 2009 J.-S. Hu, C.-Y. Chan, C.-K. Wang, M.-T. Lee, and C.-Y. Kuo [5] proposed an acoustic SLAM method based on a cross-shaped microphone array and odometry, and in 2013, N. Kallakuri, J. Even, Y. Morales, C. Ishi, and N. Hagita [6] developed a method based on a microphone array and light detection and ranging (LiDAR). The two studies above were both capable of locating the robot and mapping the environment in experiments. However, highly accurate motion reports from odometry or LiDAR are necessary for acceptable results. For example, the robot's trace measured with only motion sensors (without information from acoustic sensors) coincided perfectly with the final estimation where all sensors were involved (see experimental results in Fig. 4 of [5]). Thus, the accuracy of those methods is likely heavily dependent on the accuracy of motion sensors. Due to indoor reverberation, strong noise or even errors are common during DoA estimates. When the motion reports are inaccurate or even false, the above methods can hardly achieve acceptable results.

To facilitate robustness against false DoA estimates using motion sensors such as inertial measurement units (IMUs), C. Evers and P. A. Naylor [7], [8] developed their own acoustic SLAM with probabilistic source triangulation in 2018, and name it, appropriately, "Acoustic SLAM" (aSLAM). However, aSLAM was only tested in simulations where the measurement noise of motion reports is ideal. Specifically, simple Gaussian noise was added to the true velocity of the robot to simulate the velocity measurement. However, the noise of a typical motion sensor, such as an IMU, is mainly non-Gaussian and nonlinear. It will be demonstrated in this paper that the result of feeding the aSLAM with practical datasets is undesirable.

As mentioned before, there are occasionally false DoA estimates due to reverberation in real indoor environments. At the same time, due to the unavoidable integral error, motion sensors such as IMUs cannot provide accurate velocity logs or motion paths. Acoustic SLAM methods [5], [6], [7] that fuse DoA estimates and motion reports can be considered bearing only SLAM [9]. As the DoA estimates contain only directional information, highly accurate motion sensors are necessary to gain an acceptable overall positioning and mapping accuracy. However, low-price motion sensors, including IMUs, have evident integral error, and their noise model is nonideal. If we want to obtain acceptable SLAM results with low-accuracy motion measurements in experiments, more information from acoustic sensors is vital in addition to the DoA. In addition to bearing information, the range that denotes the distance between the source and the robot should be beneficial for acoustic SLAM, especially when false DoA estimates are common and motion reports coming from low-accuracy IMU sensors suffer from severe drifting.

Several methods have been proposed for estimating the source's distance [10], including the time difference of arrival (TDOA), deep learning, triangulation, and the direct-to-reverberant energy ratio (DRR). The TDOA is sensitive to the array size [10], and the sound source distance estimation methods based on deep learning are unable to adapt to new environments unless retraining is carried out in advance for each new environment. [11], [12]. The distance estimations with triangulation vary considerably and are still affected by the accuracy of the motion sensors [13]. The DRR method, which is based on the phenomenon of indoor sound reflections, can be used to estimate the direct source's distance from the sensor in a reverberant field. Estimates with the DRR method are insensitive to the array size and do not rely on information from motion sensors [14], so it may be suitable to estimate the range for eliminating the error of motion sensors.

In 2018, M. Strauss, P. Mordel, V. Miguet and A. Deleforge published the DREGON dataset, aiming at source localization research [15]. In this dataset, a drone with a microphone array and an IMU flies in an airtight room containing a loudspeaker. Publicly available data include the IMU data and the audio recordings during the entire flying period. Moreover, the airtight room was equipped with a motion capture system to obtain precise ground truth positions of the drone and the loudspeaker at all times. With the kind help of Dr. Deleforge, we obtained the ground truth data. The DREGON dataset contains all the information needed for the evaluation and comparison of different acoustic SLAM algorithms and will be applied here to validate our proposed method.

In this paper, a DoA-DRR acoustic SLAM (D-D SLAM) is proposed for the situation when strong noise exists in IMU motion reports. The source's distance from the robot, which is estimated using the DRR method in different time frames, is added to acoustic SLAM as a new constraint of D-D SLAM. The critical distance, which is necessary for associating the source's distance with the DRR, is estimated with an online method during the complete time period. Overcoming the deficiencies of Evers and Naylors' aSLAM [7], the proposed D-D SLAM is validated using datasets generated from simulations and real-world indoor scenes drone measurements in the DREGON dataset. Generally, this paper proposes a new methodology for acoustic SLAM using only a microphone array and IMU. The essay has been organized in the following way. Section II is the problem formulation, and Section III introduces the necessary background knowledge. Section IV derives the proposed D-D SLAM. Section V presents the setup of the simulation and experiment. Section VI shows and analyses the results. Section VII is the conclusion.

## II. PROBLEM FORMULATION

In previous acoustic SLAM research, the applied robot movement model always constrained the robot's velocity to be along the direction of orientation of the robot [5], [6], [7]. This constraint can be easily satisfied in simulation. Meanwhile, the coupling of the robot's orientation and its velocity direction simplified the mathematical deduction. However, it is only practical in a few kinds of real robots, for instance, two-wheel robots.

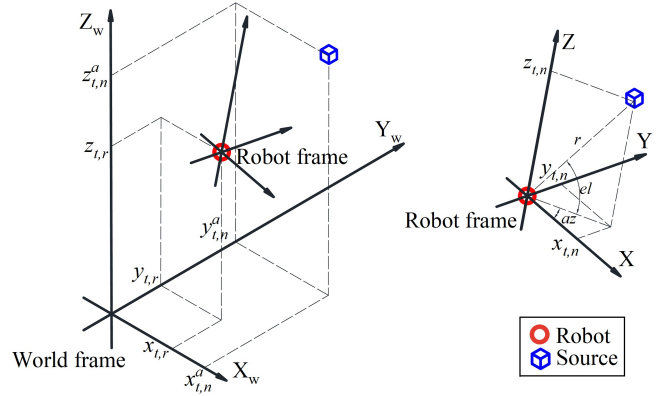

**Fig. 1.** World frame and robot frame.

For the sake of generality, a three-dimensional robot movement model is applied in the proposed method. The model decouples the direction of the robot's velocity from its orientation; thus, it is suitable for most real robots, such as drones. As depicted in Fig. 1, the robot's state at time $t$ is considered as $\mathbf{r}_t = [\mathbf{p}_t^T, \mathbf{R}_t]$, where $\mathbf{p}_t = [x_{t,r}, y_{t,r}, z_{t,r}, v_{x,t,r}, v_{y,t,r}, v_{z,t,r}]$ is the position/velocity vector and $\mathbf{R}_t$ is the rotation matrix corresponding to the robot's orientation. Symbols $az$, $el$ and $r$ in Fig. 1 denote the azimuth, elevation and radius of the sound source in the robot frame, respectively. The robot dynamics are given by:

$$\mathbf{p}_t = \mathbf{F}_t \mathbf{p}_{t-1} + d\mathbf{p}_{t-1|t} + \mathbf{v}_{t,p}, \quad \mathbf{v}_{t,\mathbf{p}} \sim \mathcal{N}(\mathbf{0}_{6\times 1}, \Sigma_{t,\mathbf{p}}) \quad (1)$$

$$\mathbf{R}_t = \zeta([\theta, \psi, \varphi]^T + \mathbf{v}_{t,\mathbf{R}}), \quad \mathbf{v}_{t,\mathbf{R}} \sim \mathcal{N}(\mathbf{0}_{3\times 1}, \Sigma_{t,\mathbf{R}}) \quad (2)$$

where $\mathbf{v}_{t,p}$ denotes unbiased Gaussian noise with covariance $\Sigma_{t,\mathbf{p}}$. $d\mathbf{p}_{t-1|t}$ is the small variation of $\mathbf{p}_t$ from time step $t-1$ to $t$, and $\zeta(\theta, \psi, \varphi)$ is a nonlinear function converting Euler angles, $[\theta, \psi, \varphi]$, to a rotation matrix. The matrix $\mathbf{F}_t$ and $d\mathbf{p}_{t-1|t}$ are given by:

$$\mathbf{F}_t = [\mathbf{I}_6] \quad (3)$$

$$d\mathbf{p}_{t-1|t} = [dx_{t-1|t}, dy_{t-1|t}, dz_{t-1|t}, dv_{x,t-1|t}, dv_{y,t-1|t}, dv_{z,t-1|t}] \quad (4)$$

where $\mathbf{I}_n$ is the $n \times n$ identity matrix. It is clear that $dx_{t-1|t}$, $dy_{t-1|t}$, and $dz_{t-1|t}$ are related to $dv_{x,t-1|t}$, $dv_{y,t-1|t}$, and $dv_{z,t-1|t}$. The method to estimate these values will be introduced in section III.

The measurements of the robot velocity $v$ and orientation $\mathbf{R}$ are defined as $\mathbf{y}_t \triangleq [y_{t,v}, y_{t,\mathbf{R}}]$ and modeled as:

$$y_{t,v} = \mathbf{h}\mathbf{p}_t + \mathbf{w}_{t,v} \quad (5)$$

$$y_{t,\mathbf{R}} = \zeta([\psi, \varphi, \theta]^T + \mathbf{w}_{t,\mathbf{R}}), \quad \mathbf{w}_{t,\mathbf{R}} \sim \mathcal{N}(\mathbf{0}_{3\times 1}, \sigma_{\mathbf{w},\mathbf{R}_t}^2) \quad (6)$$

where $\mathbf{w}_{t,v}$ is non-Gaussian noise and $\mathbf{w}_{t,\mathbf{R}}$ denotes the measurement Gaussian noise with covariance $\sigma_{\mathbf{w},\mathbf{R}_t}^2$. In addition, $\mathbf{h} \triangleq [\mathbf{0}_{3\times 3}, \mathbf{I}_3]$. In fact, robots equipped with a 9-axis IMU can provide accurate Euler angles with small errors. Therefore, D-D SLAM neglects the influence of the measurement noise on

orientation, which means that $\sigma^2_{w,R}$ is considered to be zero. Thus, (6) can be simplified to:

$$y_{t,R} = y_{t-1,R} \Delta R_{t-1|t} \quad (7)$$

where $\Delta R_{t-1|t}$ denotes the constraints of robot orientation from time step *t-1* to *t* and is introduced in Section III.

As the absolute positional state of the sound source is the landmark in mapping, the absolute state ($s^a_{t,n} \triangleq [x^a_{t,n}, y^a_{t,n}, z^a_{t,n}]^T$) of source *n* in the world frame at time step *t* is defined as [7]:

$$s^a_{t,n} = s^a_{t-1,n} + n_{t,n}, \quad n_{t,n} \sim \mathcal{N}(0_{3\times3}, Q) \quad (8)$$

where $n=1,\ldots,N_t$ is the index of $N_t$ sources, the superscript *a* stands for the world frame, and $n_{t,n}$ is the process noise with covariance $Q$. The transformation of the source's position in the world frame to that in the robot frame is given by:

$$s_{t,n} = \Gamma(s^a_{t,n} - [x_{t,r}, y_{t,r}, z_{t,r}]^T), \quad \Gamma = R_t \quad (9)$$

where $s_{t,n} \triangleq [x_{t,n}, y_{t,n}, z_{t,n}]^T$ is the positional state of the source in the robot frame and $\Gamma$ is the rotation matrix between the world frame and the robot frame.

The bearing information of the source is estimated by the DoA algorithm and is modeled as [7]:

$$\Omega_t = [\bigcup_{n=1}^{N_t} D(s_{t,n})] \cup K_t \quad (10)$$

where $D(s_{t,n})$ is the process that models the missing DoAs and estimation errors, and $K_t$ denotes the Poisson point process of $N_t$ independent and identically distributed (IID) false DoA estimates distributed uniformly over a unit sphere.

The source's distance from the sensors is also estimated solely with the DRR. The DRR estimation method is developed using interaural magnitude-squared coherence (MSC), and the source's distance is estimated by computing the discrete Fourier transform (DFT) on overlapped windowed signal frames [16]. A small sound piece is sampled at time step *t* and divided into several parts by a sliding window. In each part of the sound piece, the source's distance from the sensors is estimated. The estimation is modeled by:

$$\hat{d}^n_{t,\mu} \sim \mathcal{N}(d^n_t, R_{t,d}) \quad (11)$$

where $\mu$ denotes the windowed signal frame indices and $d^n_t$ is the true source's distance from the sensors at time step *t*. In a short time, the estimated distances $\hat{d}^n_{t,\mu}$ in different window frames at time step *t* tend to follow a nearly normal distribution that is modeled with the mean $d^n_t$ and the covariance $R_{t,d}$.

The positional state of the sources in the robot frame $s_{t,n}$ is associated with the DoA and the distances of the sources computed with the DRR. The relationship is as follows:

$$\begin{cases} [\omega_{t,m}, r_{t,m}]^T = \mathcal{G}(s_{t,n}) + e_{t,m} \\ e_{t,m} \sim \mathcal{N}(0_{3\times1}, diag(R_{t,m}, R_{t,d})) \end{cases} \quad (12)$$

where $m=1,\ldots,M_t$ denotes the index of DoA estimates, $M_t$ is the number of DoA estimates, $\omega_{t,m}=[\phi_{t,m}, \gamma_{t,m}]^T$, $\mathcal{G}(\bullet)$ is a function that is used to transform from Cartesian coordinates to spherical coordinates (azimuth $\phi_{t,m}$, elevation $\gamma_{t,m}$, and radius $r_{t,m}$), and $e_{t,m}$ denotes the measurement error with covariance which consists of DoA estimation covariance $R_{t,m}$ and distance estimation covariance $R_{t,d}$.

Estimating $r_t$ and $s^a_{t,n}$ using $\Omega_t$, $\hat{d}^n_{t,\mu}$, and $y_t$, presents more challenges than other methods.

1) The robot movement is nonuniform motion, and the velocity direction is decoupled from the robot's orientation. Therefore, the robot dynamics in this paper have more freedom of motion than in [5], [6], and [7], meaning the movement estimation becomes more complex and difficult.
2) The velocity and orientation are measured using an IMU, whose velocity measurement noise is non-Gaussian and nonlinear. This kind of noise is common in real instruments and cannot be simply removed with a traditional Kalman filter or even an extended Kalman filter (EKF) due to its strong nonlinearity.
3) The distance estimation and the DoA measurement usually intermingle with strong noise and disturbances, causing a few incorrect estimations of the sound source position, leading to no convergence.
4) The critical distance, which is essential for the estimation of the distance from the DRR, is usually calculated with the acoustic coefficients and geometry of the room. However, these parameters are unknown in our situation.

As mentioned before, new constraints, i.e., the source's distance from the sensor, are needed to overcome challenges 1 and 2. Based on a particle algorithm, a method for online estimation of the critical distance is designed to handle challenge 4, which will be introduced in Section IV. Regarding challenge 3, a filter based on a Gaussian mixture model is implemented, which will be discussed in Section IV.

III. BACKGROUND KNOWLEDGE ABOUT IMU PREINTEGRATION AND DRR

*A. IMU Preintegration*

To summarize hundreds of inertial measurements into a single relative motion constraint and update the robot states expediently, IMU preintegration [17] is used in this paper. The absolute robot state is updated by [17]:

$$R_t = R_{t-1} \Delta R_{t-1|t} \quad (13)$$

$$V_t = V_{t-1} + g\Delta t + R_t \Delta V_{t-1|t} \quad (14)$$

$$X_t = X_{t-1} + V_{t-1}\Delta t + \frac{1}{2}g\Delta t^2 + R_t \Delta X_{t|t-1} \quad (15)$$

where $X_t = [x_{t,r}, y_{t,r}, z_{t,r}]^T$, $V_t = [v_{x,t,r}, v_{y,t,r}, v_{z,t,r}]^T$, $g$ denotes the gravitational acceleration, $\Delta t$ is the time difference from time step *t-1* to *t*, and $\Delta X_{t|t-1}$ and $\Delta V_{t|t-1}$ represent the preintegrated measurements calculated with Gaussian preintegrated measurements (GPMs) [18].

*B. DRR Computing and Distance Estimator*

For distance estimation, it is useful to determine how sound is reflected in reverberant fields. The DRR is a useful ratio to estimate a source's distance [19] and can be calculated using only acoustic data. The source's distance $d(\mu)$ is estimated by:

$$d(\mu) = d_c(\sqrt{\eta_{MSC}(\mu)})^{-1} \quad (16)$$

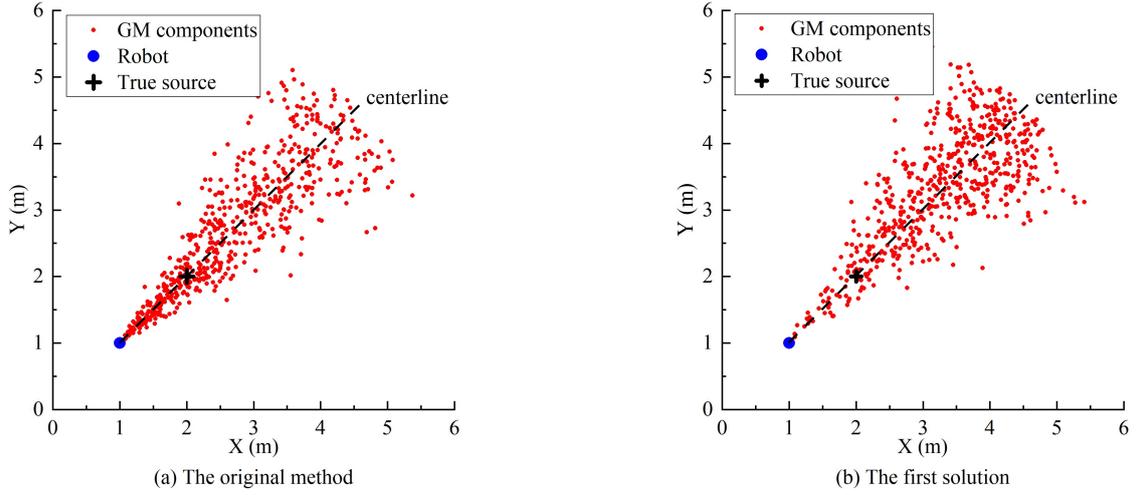

**Fig. 2.** New Gaussian mixture component distributions of (a) the original method and (b) the first solution.

where η MSC(μ) is the broadband DRR calculated with the algorithm based on interaural MSC [16] and dc is the critical distance that connects the source's distance and the DRR. The equivalent relative HRTF of the microphone array is used to apply the MSC[16] on drone data. The interaural level difference is estimated with the directivity indices of the microphone array and the interaural time difference (ITD) is calculated with the DoA result and the array shape. The coherence of the reverberant components is set as a sinc function related to the frequency index and the pairwise distance between microphones [20]. The $d_c$ is defined as [19]:

$$d_c = 0.1\sqrt{\rho_s \rho_r} \sqrt{V_R (\pi T_{60})^{-1}} \qquad (17)$$

The accuracy of the proposed D-D SLAM is based on the source's distance estimation with the DRR. A large reverberation time, which indicates a small critical distance, contributes to the improvement of the source's distance estimation with the DRR, but larger reverberation times also make the estimation of ITD and ILD more distorted, which may affect the DRR estimation [16]. If the critical distance is too small, the reverberant energy received by the microphone array is much larger than the direct energy, and thus, the ITD and ILD would be seriously affected. If the critical distance is too large, the direct energy received by the microphone array is much larger than the reverberant energy, so the acoustical field of the room is similar to the free field. The acoustical free field does not meet the subject of the proposed D-D SLAM.

For common SLAM problems, the critical distance $d_c$ is initially unknown because the source directivity indices $\rho_s$, the receiver directivity indices $\rho_r$, the room volume $V_R$ and the reverberation time $T_{60}$ cannot be measured in advance. Hence, an online method is proposed in Section IV to estimate the critical distance during SLAM exploration.

## IV. MAPPING AND LOCATING

With the fundamental theory of SLAM [21], the acoustic SLAM problem can be modeled with the SLAM posterior probability density function (PDF), which is usually factorized into two parts:

$$\begin{aligned} p(\mathbf{r}_t, \mathbf{s}_t, d_c \mid \boldsymbol{\eta}_{1:t}, \boldsymbol{\Omega}_{1:t}, \mathbf{y}_{1:t}) \\ = p(\mathbf{s}_t \mid \mathbf{r}_t, \boldsymbol{\Omega}_{1:t}) p(\mathbf{r}_t, d_c \mid \boldsymbol{\eta}_{1:t}, \boldsymbol{\Omega}_{1:t}, \mathbf{y}_{1:t}) \end{aligned} \qquad (18)$$

where $\mathbf{s}_t$ is the set of all sound source positions, $p(\mathbf{s}_t \mid \mathbf{r}_t, \boldsymbol{\Omega}_{1:t})$ is a sound source's position posterior PDF corresponding to the mapping problem in SLAM, and $p(\mathbf{r}_t, d_c \mid \boldsymbol{\eta}_{1:t}, \boldsymbol{\Omega}_{1:t}, \mathbf{y}_{1:t})$ represents the robot posterior PDF corresponding to the locating problem in SLAM. The critical distance $d_c$ is a parameter to be estimated. $\boldsymbol{\eta}_{1:t}$ denotes the DRR estimates from the beginning to time step $t$ that are computed with the method based on interaural MSC [16].

*A. Mapping*

The sound sources are considered landmarks on the map, so the focus of the mapping procedure is the estimation of the source positions. To solve existing problems, including false DoA estimates, C. Evers and P. A. Naylor [7] proposed a mapping method based on probabilistic source triangulation and a random finite set. Their method performs well when the robot keeps moving and receiving a signal from the sound source. However, when the positions of both the robot and the sound source remain stationary, mapping mistakes may occur during the emerging process, which is often used to limit the number of Gaussian mixture (GM) components. In this situation, the robot's position and DoA estimates do not change, resulting in the repeated emergence of the same new GM components located in the same zone. As the new GM components satisfy both the wrapped Gaussian distribution in angle and the uniform distribution in the radial direction [7], the component's distribution density in the sector region is uneven. The distance between any two GM components $l$ can be calculated by:

$$l = \sqrt{2r^2 + \Delta r^2 + 2r\Delta r(1-\cos(\Delta\gamma))} \qquad (19)$$

where $r$ is one of the component's distances from the robot, and $\Delta r$ and $\Delta\gamma$ are the difference of the two components in

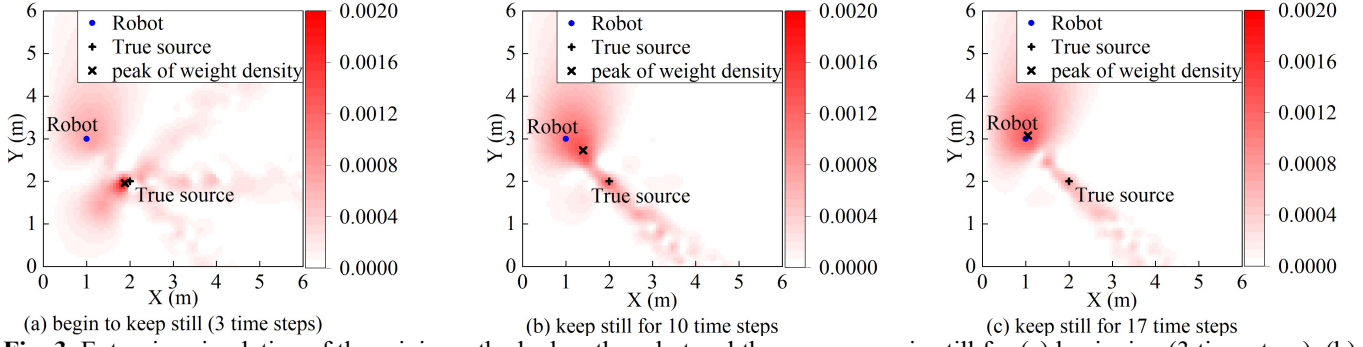

(a) begin to keep still (3 time steps)  (b) keep still for 10 time steps  (c) keep still for 17 time steps

**Fig. 3.** Extensive simulation of the origin method when the robot and the source remain still for (a) beginning (3 time steps), (b) 10 time steps, and (c) 15 time steps.

radius and angle, respectively. With the same $\Delta r$ and $\Delta \gamma$, the shorter the radius distance of the GM component is, the shorter the distance between different GM components is. Thus, the component density in the region closer to the robot is higher than others. Fig. 2(a) reveals this false trend (which can also be found in Fig. 3 of [7]).

Before the clustering of GM components for source estimation, mixture reduction is usually applied to limit the number of components through merging. The criterion equation of merging components is given by [22]:

$$(m_k^i - m_k^j)^T (P_k^i)^{-1} (m_k^i - m_k^j) \leq U, \quad i=1,\ldots,J_k \quad (20)$$

where $J_k$ is the number of GM components, $j$ is the index of the component with the maximum weight, $m_k^i$ denotes the position of the GM component in the robot frame, $P_k^i$ is the covariance of the GM component's position and $U$ is the threshold of merging.

With the criterion equation (20), it is found that the component merging is relevant to the covariance and the range between components. The new components are created with the same covariance and weight. Therefore, when they have a denser distribution (closer to the robot), they are more likely to be merged together, forming merged components with larger weights. When both the sound source and the robot remain stationary, new components will be created repeatedly in the same zone, and the distribution of weights (after merging) will become increasingly uneven over time. The area closer to the robot will receive much larger weights compared with that far from the robot. This trend has a negative influence on GM component clustering and may finally lead to a false estimation of the sound source position, indicating that it is much closer than the true value. Here, a simulation test is conducted to illustrate the problem mentioned above, while two solutions will also be introduced in this section. The simulation test setup is basically the same as that of part C in Section IV of [7], while an extra time segment is added when the robot remains stationary for several time steps.

Fig. 3 is a heatmap of the source weight density calculated using the origin method [7] in the simulation setup with an extra time segment. A higher saturation level of red indicates a higher weight density. As shown in Fig. 3(a), just after the robot moved for three time steps (same as that in the original simulation [7]), the GM components were clustered into the place close to the true source where the probability density peak is located. However, as shown in Fig. 3(b) and (c), the longer the robot remains stationary, the greater the peak of the probability density deviates toward the robot. The same trend can be found with the weighted centroid of the GM components. This deviation causes an incorrect estimation of the source position, leading to incorrect mapping.

To fix this problem, two solutions are developed. The first solution is to improve the creation method of new components. To avoid heterogeneous density of the new GM component in the direction of the DoA, a new method is developed instead of the original distribution [7]:

$$\hat{r}_{t,m}^{(j)} \sim r_{min} \sqrt{(\mathcal{U}(1, \frac{r_{max}^2}{r_{min}^2}))} \quad (21)$$

where $\hat{r}_{t,m}^{(j)}$ denotes the robot's distance from GM new component $j$, $r_{min}$ and $r_{max}$ are the practical lower and upper limits of $\hat{r}_{t,m}^{(j)}$, respectively, and $\mathcal{U}(a, b)$ is the uniform distribution between $a$ and $b$. Equation (21) ensures that new GM components will be created uniformly in a sector-shaped region, as shown in Fig. 2(b). Thus, the weighted centroid of the GM components will not be shifted toward the robot. Instead, the cluster of all GM components will be shifted to the closest point on the centerline, which is more likely the position of the true source.

The second solution is the use of keyframes. When the robot and source remain stationary, new GM components are always created in the same region so that the new GM components are merged into a false component with a large weight. The weight of the false component increases over time, and the weighted centroid of the components is eventually shifted toward the robot. If the creation and merging of GM components stop when the robot and source remain stationary, a false component with a large weight is avoided. The basic idea is that a keyframe factor is calculated for evaluating the difference between the current frame and the last keyframe:

$$T = (\hat{T}_t^1 > T_{KF}^1) \| (\hat{T}_t^2 > T_{KF}^2) \| (\hat{T}_t^3 > T_{KF}^3) \quad (22)$$

$$\begin{cases} \hat{T}_t^1 = \|(x_{t,r}, y_{t,r}, z_{t,r})^T - (x_{KF,r}, y_{KF,r}, z_{KF,r})^T\| \\ \hat{T}_t^2 = \|\Omega_t - \Omega_{KF}\| \\ \hat{T}_t^3 = \|\zeta^{-1}(R_t R_{KF}^T)\| \end{cases} \quad (23)$$

where the subscript $KF$ denotes a keyframe, $T_{KF}^k$ for $k=1,2,3$ denotes the given threshold of the keyframe, $\|\cdot\|$ is the two-

norm, $\zeta^{-1}(\cdot)$ is a function converting a rotation matrix to Euler angles, and || indicates the OR operation. When the robot and the source remain stationary, $\hat{T}_t^k$ will be less than $T_{KF}^k$, making $T$ false. When $T$ is false, the current frame in time step $t$ is not a keyframe. In contrast, when the robot moves far enough, $\hat{T}_t^k$ becomes larger than $T_{KF}^k$, making $T$ true. Thus, the current frame is considered a new keyframe, and the state of the keyframe is updated with the following equations:

$$\begin{cases} (x_{KF,r}, y_{KF,r}, z_{KF,r})^T = (x_{t,r}, y_{t,r}, z_{t,r})^T \\ \Omega_{KF} = \Omega_t \\ R_{KF} = R_t \end{cases} \quad (24)$$

When GM components are only created and merged in keyframes, a false merged component with a large weight is avoided. For the purpose of reducing computation and complexity, the second solution is applied in the proposed D-D SLAM.

To eliminate accidental error, a limiting filter is applied to the clustering of GM components. The limiting filter is modeled as:

$$LF(s_t) = \begin{cases} s_t, & \|s_t - s_{t-1}\| \leq s_{LF} \\ s_{t-1}, & \|s_t - s_{t-1}\| > s_{LF} \end{cases} \quad (25)$$

where $s_{LF}$ denotes the threshold value of the limiting filter, which is related to the maximal variation of the source position between time steps $t$-1 and $t$.

Using the probabilistic source triangulation in [7] and the keyframe method mentioned above, the source posterior PDF $p(\mathbf{s}_t | \mathbf{r}_t, \mathbf{\Omega}_{1:t})$ can be estimated using the evidence of the DoA estimation $\mathcal{L}(\mathbf{\Omega}_t | \mathbf{r}_t, d_c)$. The implementation is given by:

$$\mathcal{L}(\mathbf{\Omega}_t | \mathbf{r}_t, d_c) \triangleq e^{-N_{t,c} - p_d N_{t|t-1}} \prod_{m=1}^{M_t} \ell(\boldsymbol{\omega}_{t,m} | \mathbf{r}_t, d_c) \quad (26)$$

where $p_d$ denotes the probability of detection, $N_{t,c}$ is the number of false alarms, $N_{t|t-1}$ stands for the quantity of predicted features [7], and $\boldsymbol{\omega}_{t,m}$ is the DoA estimation computed from the GM components and the position of the robot.

### B. Locating

This section proposes a method to estimate the robot location by fusing the IMU measurements and the DoA and DRR estimates. Additionally, the critical distance is estimated online. Because the relationship between the state of the robot and the sensor measurements is nonlinear and complicated, it is difficult to estimate the critical distance $d_c$ and the state of the robot $\mathbf{r}_t$ directly from just the IMU measurements and the DoA and DRR estimates. Particle filters [23] are usually applied to model the robot posterior PDF of this nonlinear problem. With this method, the robot posterior PDF $p(\mathbf{r}_t, d_c | \boldsymbol{\eta}_{1:t}, \boldsymbol{\Omega}_{1:t}, \mathbf{y}_{1:t})$ is modeled as:

$$p(\mathbf{r}_t, d_c | \boldsymbol{\eta}_{1:t}, \boldsymbol{\Omega}_{1:t}, \mathbf{y}_{1:t}) \approx \sum_{i=1}^{I} \alpha_t^i \delta_{\hat{\mathbf{r}}_t^i, \hat{d}_c^i}(\mathbf{r}_t, d_c) \quad (27)$$

where $I$ is the number of particles, $\alpha_t^i$ denotes the weight of a particle and $\delta_{\hat{\mathbf{r}}_t^i, \hat{d}_c^i}(\mathbf{r}_t, d_c)$ is the Dirac-delta function centered at $\mathbf{r}_t$, $d_c$ and evaluated at $\hat{\mathbf{r}}_t^i$, $\hat{d}_c^i$. Because the state of the robot and the critical distance are estimated at the same time, there is a large amount of uncertainty so that large quantities of particles are required to model the robot posterior PDF.

Using (16), the source's distance from robot $\hat{d}$ can be computed with the DRR if the critical distance is known. As the source's distance is helpful in the estimation of the robot position, it is easier to estimate the state of the robot when the critical distance is known. To simplify the robot posterior PDF, the marginalization [24] is adopted to decouple the posterior PDF into that of $d_c$ and $\mathbf{r}_t$ separately, i.e., the robot posterior PDF $p(\mathbf{r}_t, d_c | \boldsymbol{\eta}_{1:t}, \boldsymbol{\Omega}_{1:t}, \mathbf{y}_{1:t})$ is factorized into two parts:

$$\begin{aligned} & p(\mathbf{r}_t, d_c | \boldsymbol{\eta}_{1:t}, \boldsymbol{\Omega}_{1:t}, \mathbf{y}_{1:t}) \\ & = p(d_c | \boldsymbol{\eta}_{1:t}, \boldsymbol{\Omega}_{1:t}, \mathbf{y}_{1:t}) p(\mathbf{r}_t | d_c, \boldsymbol{\eta}_{1:t}, \boldsymbol{\Omega}_{1:t}, \mathbf{y}_{1:t}) \end{aligned} \quad (28)$$

Using (15), the state of the robot can be computed from the IMU measures. Meanwhile, the source's distance from the robot computed from a given $d_c$, $\boldsymbol{\eta}_{1:t}$ using (16), and DoA estimates can be used to correct the state of the robot. Therefore, this paper takes the estimation of the robot's state $\mathbf{r}_t$ as a nonlinear substructure [24] of the estimation of critical distance $d_c$. Thus, for each $d_c$ particle, there is a substructure corresponding to the robot state and the source position.

The critical distance is assumed to be within a certain range $\hat{d}_c \in [d_c^{min}, d_c^{max}]$ and obeys a uniform distribution:

$$\hat{d}_c^i \sim \mathcal{U}(d_c^{min}, d_c^{max}), \quad i = 1, ..., I \quad (29)$$

where $d_c^{min}$ and $d_c^{max}$ are the minimum and maximum of the critical distance, respectively, and $I$ is the number of particles. A critical distance particle is drawn from the uniform distribution. For each critical distance particle and each source, the estimations of the source's distance at time step $t$ can be calculated in the following formula according to (16):

$$\hat{d}_{t,m,\mu}^i = \hat{d}_c^i (\sqrt{\eta_{t,m,\mu}})^{-1}, \quad \mu = 1, ..., B \quad (30)$$

where $B$ denotes the number of windowed signal frames and $m=1,...,M_t$. To eliminate accidental and obvious error, a rate limiter [25] is applied in $\hat{d}_{t,m,\mu}^i$ to limit the rate of change with a threshold value that is related to the maximum robot velocity.

Theoretically, the state of the robot can be computed using iterative calculation of (13), (14) and (15) as long as the initial value of the robot's state is given. In fact, a 9-axis acceleration gyroscope sensor can provide an accurate rotation matrix $\mathbf{R}_t$ that can be used to accurately describe the attitude of the robot. Therefore, $\mathbf{R}_t$ is computed directly from IMU measurements and the GPMs in this paper. However, the velocity and position of the robot cannot be computed using only the IMU measurements due to unacceptable integral error. Therefore, the DoA estimates and DRR are used to eliminate those errors by fusing all measures with an EKF [26]. For weakly nonlinear systems, an EKF has better performance than a Kalman filter. Therefore, it is suitable for the estimation of the weakly nonlinear state of the robot in D-D SLAM. Using (13), (14) and (15), the implementation of the EKF is given by:

$$\mathbf{X}_t^{\prime i} = \mathbf{X}_{t-1}^i + \mathbf{V}_{t-1}^i \Delta t + \frac{1}{2} \mathbf{g} \Delta t^2 + \mathbf{R}_t \, d\mathbf{X}_{t-1|t} \quad (31)$$

where $\mathbf{X}_t^{\prime i}$ denotes the prediction of the robot's position in the EKF. The rotation matrix $\mathbf{R}_t$ of the robot is determined from (13). The preintegrated measurement of position $d\mathbf{X}_{t-1|t}$ is given

with the GPMs [18]. The prediction of covariance in the EKF is given by:

$$Cov_{t,m}^{'i} = F_t Cov_{t,m}^{i} F_t^T + Q_{X_t}^{i} \quad (32)$$

where $Cov_{t,m}^{i}$ denotes the covariance of the EKF, $F_t$ is given by (3) and $Q_{X_t}^{i}$ is the process non-Gaussian noise of the robot position. $Q_{X_t}^{i}$ is associated with the process noise of the robot's velocity, according to (31) with the variance-covariance propagation law:

$$Q_{V_t}^{i} = Q_{V_{t-1}}^{i} + dQ_{V_{t-1|t}} \quad (33)$$

$$Q_{X_t}^{i} = Q_{X_{t-1}}^{i} + \Delta t\, Q_{V_{t-1}}^{i} \Delta t + dQ_{X_{t-1|t}} \quad (34)$$

where $dQ_{X_{t-1|t}}$ and $dQ_{V_{t-1|t}}$ denote the variances of the GPMs of the robot position and velocity, respectively. For each source, the observational equation in the EKF is determined by:

$$[\hat{\omega}_{t,m}^{i}, \hat{r}_{t,m}^{i}]^T = \mathcal{G}(R_t(X_t^{'i} - \hat{s}_{t,m}^{i})) \quad (35)$$

where $m=1,\ldots,M_t$ and $\mathcal{G}(\cdot)$ is the Cartesian-to-spherical transformation. According to the mapping procedure (Section IV part A), the estimation of source $\hat{s}_{t,m}^{i}$ is given using GM component clustering. Hence, the Kalman gain and the correction of Kalman gain are given by:

$$K_{t,m}^{i} = Cov_{t,m}^{'i} (H_{t,m}^{i})^T (H_{t,m}^{i} Cov_{t,m}^{'i} (H_{t,m}^{i})^T + R_{EKF})^{-1} \quad (36)$$

$$\hat{X}_{t,m,\mu}^{i} = X_t^{'i} + K_{t,m}^{i}([\Omega_{t,m}, \hat{d}_{t,m,\mu}^{i}]^T - [\hat{\omega}_{t,m}^{i}, \hat{r}_{t,m}^{i}]^T) \quad (37)$$

$$\hat{V}_{t,m,\mu}^{i} = (\hat{X}_{t,m,\mu}^{i} - X_{t-1}^{i})/dt \quad (38)$$

$$Cov_{t,m}^{i} = (I_3 - K_{t,m}^{i} H_{t,m}^{i}) Cov_{t,m}^{'i} \quad (39)$$

where $H$ is the Jacobian matrix of the observational equation, $\hat{V}_{t,m,\mu}^{i}$ is the velocity of the robot corresponding to each output of the EKF, $dt$ denotes the time difference between two adjacent keyframes, which is different from $\Delta t$, and $R_{EKF}$ is the measurement noise, which is assumed to be known. The output of the EKF, $\hat{X}_{t,m,\mu}^{i}$, corresponds to each source's distance $\hat{d}_{t,m,\mu}^{i}$. As the estimation of the robot's position is nonlinear and the rotation matrix $R_t$ is computed directly from IMU measurements, the Gaussian mixture model (GMM) is applied to model the robot posterior PDF $p(X_t | d_c^i, \eta_{1:t}, \Omega_{1:t}, y_{1:t})$:

$$p(X_t | d_c^i, \eta_{1:t}, \Omega_{1:t}, y_{1:t})$$
$$= \sum_{M_t} \sum_{B}^{\mu} w_{t,m,\mu}^{j} \mathcal{N}(X_t | \hat{X}_{t,m,\mu}^{i}, Cov_{t,m}^{i}) \quad (40)$$

where $w_{t,m,\mu}^{j}$ denotes the weights of the GM components in the windowed signal frame indices $\mu$ of time step $t$ and is given by:

$$w_{t,m,\mu}^{j} = \mathcal{N}(\hat{X}_{t,m,\mu}^{i} | X_t^{'i}, Q_{X_t}^{i}) \quad (41)$$

It is clear that the smaller the difference between the outputs of the EKF and GPMs is, the larger the component weight $w_{t,m,\mu}^{i}$ is. For the GMM, the estimation of the robot position, the robot velocity, the source's distance from the robot, and the covariance in the EKF are computed with the weighted average method

$$X_t^{i} = \sum_{M_t} \sum_{B}^{\mu} w_{t,m,\mu}^{j} \hat{X}_{t,m,\mu}^{i} \quad (42)$$

$$V_t^{i} = \sum_{M_t} \sum_{B}^{\mu} w_{t,m,\mu}^{j} \hat{V}_{t,m,\mu}^{i} \quad (43)$$

$$d_{t,m}^{i} = \| X_t^{i} - \hat{s}_{t,m} \| \quad (44)$$

$$Cov_{t,m} = \sum_{M_t} \sum_{B}^{\mu} w_{t,m,\mu}^{i} Cov_{t,\mu}^{i} \quad (45)$$

The weight of each critical distance and robot position particle is given by:

$$\alpha_t^{i} = \mathcal{N}(X_t^{i} | X_t^{'i}, Q_{X_t}^{i}) \quad (46)$$

Equation (46) reflects the coincidence degree between the robot position corresponding to each critical distance particle and the GPM measurements. The GPMs of the IMU are used to evaluate each critical distance particle. The weights of the critical distance particle that fits the GPM measurements well will become larger. The final estimation of the robot and source position will be in the best interests of all measurements.

### C. Posterior PDF of the D-D SLAM

This section provides the implementation of evaluating the SLAM posterior PDF. Because a rotation matrix can describe the attitude of the robot accurately with the GPMs of the IMU data and the source triangulation does not depend on velocity, (18) is reduced to:

$$p(X_t, s_t, d_c | \eta_{1:t}, \Omega_{1:t}, y_{1:t})$$
$$= p(s_t | X_t, \Omega_{1:t}) p(X_t, d_c | \eta_{1:t}, \Omega_{1:t}, y_{1:t}) \quad (47)$$

According to the Bayes rule, using (27) and (26), the posterior PDF $p(X_t, d_c | \eta_{1:t}, \Omega_{1:t}, y_{1:t})$ is given by:

$$p(X_t, d_c | \eta_{1:t}, \Omega_{1:t}, y_{1:t})$$
$$= \frac{\sum_{i=1}^{I} \alpha_t^{i} \mathcal{L}(\Omega_t | \hat{X}_t, \hat{d}_c) \delta_{\hat{X}_t, \hat{d}_c^{i}}(X_t, d_c)}{\sum_{j=1}^{I} \alpha_t^{j} \mathcal{L}(\Omega_t | \hat{X}_t, \hat{d}_c)} \quad (48)$$

Using (48) in (47), the SLAM posterior PDF is reduced to:

$$p(X_t, s_t, d_c | \eta_{1:t}, \Omega_{1:t}, y_{1:t})$$
$$= \sum_{i=1}^{I} \beta_t^{i} \delta_{\hat{X}_t, \hat{d}_c^{i}}(X_t, d_c) \, p(s_t | X_t, \Omega_{1:t}) \quad (49)$$

$$\beta_t^{i} = \frac{\alpha_t^{i} \mathcal{L}(\Omega_t | \hat{X}_t, \hat{d}_c)}{\sum_{j=1}^{I} \alpha_t^{j} \mathcal{L}(\Omega_t | \hat{X}_t, \hat{d}_c)} \quad (50)$$

where $\beta_t^{i}$ is the weight and is used to evaluate the particle and estimate the position of the robot and the source. The D-D SLAM is summarized in pseudocode (see Table I).

Each particle is now evaluated by the evidence of mapping (26) and the weight of locating (46). When the velocity of the robot is updated with (43), first-order recursive temporal smoothing is applied to smooth the speed to minimize the jitter of the velocity:

$$V_t^{i} = a_p V_{t-1}^{i} + (1 - a_p) \sum_{M_t} \sum_{B}^{\mu} w_{t,m,\mu}^{i} \hat{V}_{t,m,\mu}^{i} \quad (51)$$

where $a_p$ is a smoothing parameter. In the procedure "Update particle state", the estimation of the robot and the source at

TABLE I

**Algorithm 1: D-D SLAM**

1 **Data:** DoAs $\Omega_t$, DRR $\eta_t$, IMU Measure $\mathbf{y}_t$
2 **Result:** The state of robot and source $\mathbf{r}_t$, $\mathbf{s}_t$
3 **for** $i=1, ..., I$ **do**
4     Sample $\hat{d}_c^i$ using (29);
5     Compute $\hat{\mathbf{r}}_t^i$ using (13)(14)(15);
6     Compute KeyFrame factor using (22)(23);
7     **if** *KeyFrame* **then**
8        Compute $Q_{\mathbf{X}_t}^i$, $Cov_{t,m}^{\prime i}$ using (34)(33)(32);
9        **for** $m=1, ..., M_t$ **do**
10           Predict $\hat{\omega}_{t,m}^i$, $\hat{r}_{t,m}^i$ using (35);
11           Compute $\mathbf{K}_{t,m}^i$ using (36);
12           **for** $\mu=1, ..., B$ **do**
13              Evaluate $\hat{\mathbf{X}}_{t,m,\mu}^i$, $\hat{\mathbf{V}}_{t,m,\mu}^i$ using (37)(38);
14              Compute $w_{t,m,\mu}^i$ using (41);
15           **end**
16        **end**
17        Update $Cov_{t,m}$ using (45);
18        Update $\mathbf{X}_t^i, \mathbf{V}_t^i$ using (42)(43);
19        Evaluate $\alpha_t^i$ using (46);
20        Compute $\mathbf{s}_t^i$ using the improved mapping method in Section IV;
21        Evaluate $\mathcal{L}(\Omega_t \mid \mathbf{r}_t, d_c)$ using (26);
22     **else**
23        Update $\hat{\mathbf{r}}_t^i$, $Q_{\mathbf{X}_t}$ using (13)(14)(15)(34)(33);
24     **end**
25     GM reduction [27];
26     Evaluate $\beta_t^i$ using (50);
27     Update particle state;
28 **end**
29 Resampling [28];

time step $t$ is computed with the weighted mean method.

## V. SIMULATION AND EXPERIMENT SETUP

*A. Simulation Setup*

This simulation is designed to compare the performances of D-D SLAM and aSLAM [7] in a simulation room. In the simulation, the feasibility of the proposed online estimation algorithms for the critical distance used in D-D SLAM is also illustrated. The simulation room is similar to that in [7], i.e., a sealed 6 m×6 m×3 m room. A continuous signal source is placed in the center of the room (3 m, 3 m, 1.5 m), and the reverberation time is set to 0.15 s. The robot moves in the direction of a random orientation, similar to the movement in [7]. The magnitude of the velocity is set to 2 m/s with reference to the famous robot, TurtleBot2.

The trajectory and IMU data are simulated using Robot Operating System (ROS) and Gazebo. A TurtleBot2 equipped with a microphone array and IMU is set in the room and moves randomly. The IMU data that are simulated by ROS contain only the numerical error. For vraisemblance, two types of Gaussian white noise are added artificially to the output data of the accelerometer and the gyroscope with variances of $1\times10^{-3}$ and $1\times10^{-2}$, respectively, similar to that of the familiar low-cost IMU MPU6050. Therefore, the simulated IMU data contain both the nonlinear numerical error and the Gaussian measurement noise. The true trajectory data are recorded as the ground truth.

A sound record with eight channels is generated according to the image source method [29], [30] using a room impulse response (RIR) simulator. The sample frequency is set to 16000 Hz, and the microphone array shape is the same as in the DREGON dataset [15] mentioned above. The microphone array is fixed on the robot, and the transformation matrix between their positions and orientations is constant.

The DoA method based on SRP-PHAT [31] is applied to estimate the direction of the source, and the GPMs are applied to provide the observed values of the velocity, position and orientation. With SRP-PHAT, the error of the DoA estimates is less than 2 degrees. The DRR computation is introduced in Section III.

In [7], the measured velocity is simulated by adding Gaussian noise directly to the true velocity, which is different from the true IMU model. However, the velocity computed using IMU integration contains accumulated error, which is nonlinear and cannot be eliminated simply by a Kalman filter. As a result, the accuracy of aSLAM with true IMU data will decrease over time. For comparison, three groups are developed in this simulation. Group I: aSLAM with motion reports, which is simulated by adding Gaussian noise to the true velocity directly, i.e., the same as in [7]. The noise signal is unbiased, and the root mean squared error (RMSE) of the noise signal is 0.75 m/s for velocity and 5 deg for azimuth. Group II: aSLAM with motion reports, which is computed with GPMs on the simulated IMU data that contain both nonlinear numerical error and Gaussian measurement noise. Group III: the proposed D-D SLAM with the same simulated IMU data as that in Group II. In all groups, the same DoA measurements are used, the robot starts in the same place, the number of particles is set to 10, the covariance of the DoAs is set to 2 deg, and the covariance of the source's distance from the robot is set to 0.35 m. In this simulation, the time step, i.e., the time difference between nearby frames, is 1 s, and the total simulation time is 105 s (105 data frames).

*B. Experiment Setup*

To evaluate the effects of the algorithms on real indoor data, the DREGON dataset [15], which contains real data sampled with a drone flying in a real room, is applied. In the DREGON dataset, a MikroKopter unmanned aerial vehicle (UAV) equipped with a microphone array and IMU flew indoors. In this paper, the Free Flight-White Noise Source at High Volume case in the DREGON dataset is used. A continuous white noise sound source was placed in the room and sampled by an array of eight microphones (8SoundsUSB and ManyEars). The motion reports were measured with the onboard IMU (which is integrated in FlightCtrl 2.5). Meanwhile, the positions and orientations of the UAV and the source were recorded precisely by a 12-camera Vicon motion capture system and considered the ground truth. According to the dataset, the speed of the UAV is no faster than 1 m/s, and the flight contains hovering, a rectangle, spin, up, and down.

It is clear that the magnitude of the UAV velocity is variable and that the direction of the UAV velocity is independent of the orientation of the UAV. To test aSLAM on the DREGON dataset, the robot dynamics must be updated for the flight patterns of the UAV. For comparison, two experimental groups were used. Group IV: aSLAM [7] with updated observer dynamics. Group V: the proposed D-D SLAM. In both groups, the DoA estimates and motion reports are computed using the same method as in the simulation, and the particle number is set to 10. The UAV started in the same place and with the same posture. On the real indoor dataset, the elevation search boundaries are limited in [-90°, 20°] to avoid the drone noise's influence, which has an elevation angle that is mostly 60°. The error of the DoAs estimated with SRP-PHAT is also less than 2 degrees. Then, the covariance of the DoA estimates is set to 2 degrees. The covariance of the source's distance estimation from the robot is set to 0.35 m. In this experiment, the time step, i.e., the time difference between nearby frames is 0.0464 s, and the total simulation time is 46 s (993 data frames).

*C. Performance Metric*

To quantitatively analyze the accuracy, the error between the positional estimation and ground truth is evaluated with the Euclidean distance. The Euclidean distance is calculated by:

$$d(\mathbf{X}_t^{gt}, \mathbf{X}_t^{est}) = \| \mathbf{X}_t^{gt} - \mathbf{X}_t^{est} \| \quad (52)$$

where the superscripts *gt* and *est* denote the ground truth and estimation, respectively, and $\mathbf{X}_t^{(\cdot)}$ is the position of the robot or the source.

## VI. THE RESULTS

*A. Simulation Results*

The results of Groups I, II and III in the simulation are shown in Fig. 4. As shown in Fig. 4 with the orange dashed-dotted line and solid circles, the trajectory and source position estimations of aSLAM with the true speed, which only contains Gaussian noise similar to that in [7], reach the expected effect, meeting the mean accuracies of 0.136 m and 0.14 m for estimations of the trajectory and source position, respectively.

However, when the simulated IMU data that contain nonlinear numerical error and Gaussian measurement noise are applied, aSLAM has poor performance. As shown by the blue dashed line in Fig. 4, the trajectory estimation deviation increases with time. Within 29 s (29 time steps), the estimated position of the robot starts to be outside the room. The corresponding source position estimations are nonconvergent and sometimes even out of the room, so they cannot be marked as static points in Fig. 4. In contrast, the proposed D-D SLAM with the same simulated IMU data performs well. As shown in Fig. 4, the trajectory estimation of D-D SLAM in the red dotted line basically matches the ground truth, and the source position estimation in the red solid triangles is always close to the true source's position. During the whole process of 105 s (105 time steps), the estimation of the trajectory and source position with the proposed D-D SLAM is convergent and stable.

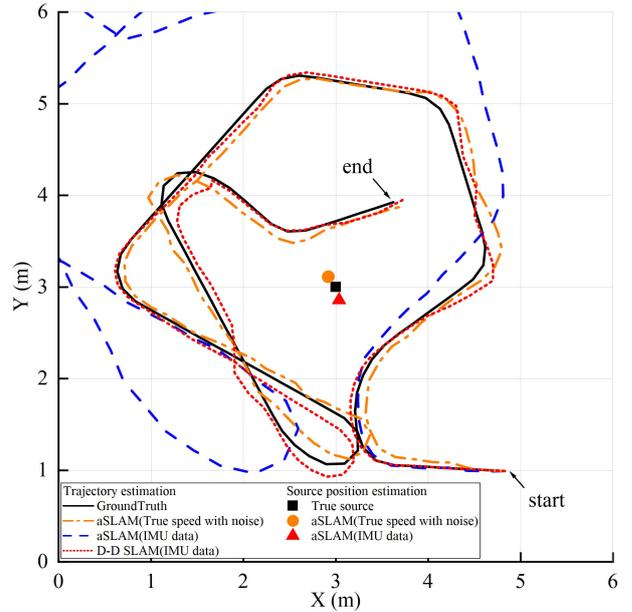

**Fig. 4.** Trajectory estimation on simulations for aSLAM with true speed plus noise (orange dash-dotted line), aSLAM with IMU data (blue dashed line) and D-D SLAM with IMU data (red dotted line).

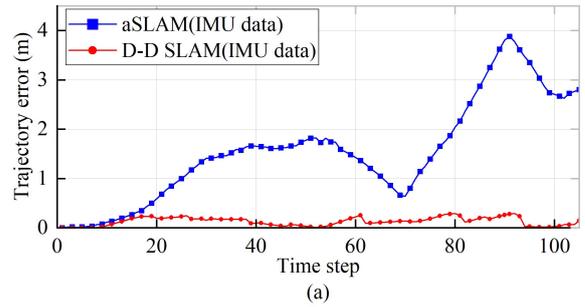

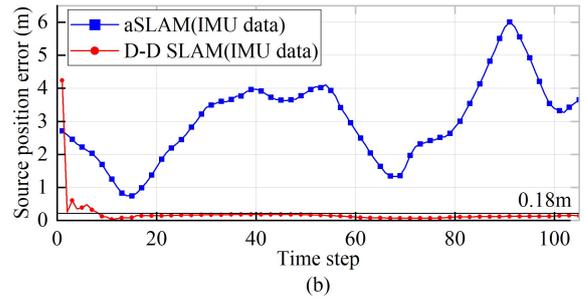

**Fig. 5.** Simulation results for (a) the trajectory error and (b) the source position error over time steps.

The poor performance of aSLAM with IMU data implies that additional constraints are needed in this situation. As the nonlinear noise is due to the integration error of the IMU, the DRR is used to compute the source's distance from the robot, which is applied as a new constraint. The result shown in Fig. 4 demonstrates the different performances under the same conditions between aSLAM and the proposed D-D SLAM, proving the effectiveness of D-D SLAM. The use of the DRR for source distance estimations, which is considered an additional constraint, is key to eliminating the nonlinear error of IMU measurements.

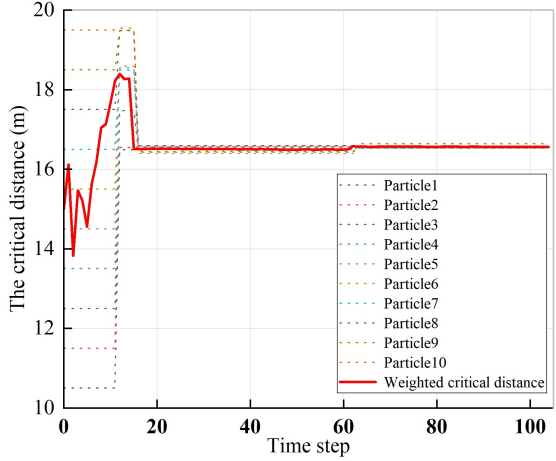

**Fig. 6.** The state of different critical distance particles and the weighted critical distance with time in the simulation.

Quantitative analysis of the accuracy using the Euler distance is illustrated in Fig. 5. The result of aSLAM is drawn as a blue line with solid squares, while that of the proposed D-D SLAM is drawn as a red line with solid circles.

For aSLAM with IMU data, the trajectory error reaches a maximum of 3.9 m at 91 s (91 time steps), and its mean value is 1.55 m, as shown in Fig. 5(a). Moreover, the overall trend of its trajectory error increases over time. With the random initial source position, the corresponding source position estimation error is unstable, as shown in Fig. 5(b). The error also reaches a maximum of 6.0 m at 91 s (91 time steps), and its average is 3.0 m. Within a room of 6 m×6 m×3 m, the trajectory and source position errors are unacceptable.

For the proposed D-D SLAM with IMU data, the maximum trajectory error is less than 0.29 m, and the mean value of the trajectory error is 0.127 m. Meanwhile, the source position error is less than 0.18 m just after 9 s (9 time steps), even though the source position error at the beginning is quite large due to the random initial source position. Compared with aSLAM, the source position error of D-D SLAM converges faster even with the larger initial value of the source position error. These results show that the proposed D-D SLAM has better performance using the same simulated IMU measurement, which proves that D-D SLAM has stronger robustness for the nonlinear noise of motion reports.

The good performance of the proposed D-D SLAM relies on reliable source distance estimates which are based on the accurate estimate of the critical distance $d_c$. However, $d_c$ is initially unknown. With the online estimation method in this paper, $d_c$ is calculated during the SLAM process. Fig. 6 is the result of $d_c$ at different time steps, where the different dashed lines represent values of different $d_c$ particles and the solid line represents their weighted mean. The mean of $d_c$ converges rapidly to a stable value in less than 18 s (18 time steps). The results show that the proposed online estimation method can obtain a convincing critical distance, even though the directivity of the source and receiver, the room size and the reverberation time are unknown.

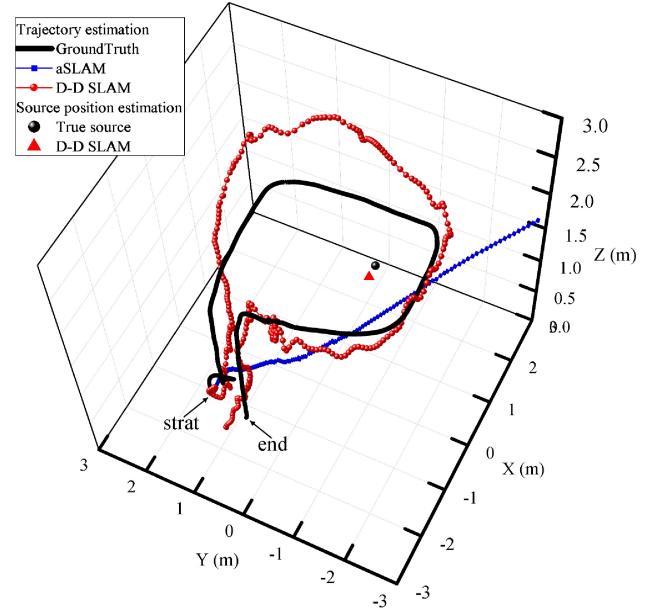

**Fig. 7.** The estimation of trajectory and the source position using aSLAM and D-D SLAM.

*B. Experimental Results*

Fig. 7 shows the estimated trajectory and source position using D-D SLAM and aSLAM for real indoor datasets [15]. Even worse than that in simulation Group II, the estimated trajectory of aSLAM in the experiment goes outside the boundary of the experiment room after 18.7 s (402 time steps) and never returns during the remainder of the experiment. Therefore, it is not fully shown in Fig. 7. The gradual divergence of the estimated trajectory is because of the shifting integrated velocity from the IMU measurements. Due to the incremental error of trajectory estimation, the source estimation of aSLAM is also unstable and out of the room, so it cannot be marked in Fig. 7. In contrast, the estimated trajectory of the proposed D-D SLAM always converges to a small neighborhood of the ground truth, as shown in Fig. 7. Meanwhile, the estimation of the source position also converges to the true source position.

Fig. 8 illustrates the estimation errors of the trajectory and source position. For aSLAM with real indoor data, the overall trajectory error tends to increase during the complete SLAM process, which is unacceptable. The source position error has the same trend as that of the trajectory error.

For the proposed D-D SLAM, the maximum trajectory error reaches 1.18 m at 16.1 s (347 time steps), and its mean value is 0.48 m. In the meantime, the maximum error of the source position estimates of 0.66 m occurs at the beginning because the initial source position is random, and the error quickly converges to the stable value of 0.25 m after 2.8 s (60 time steps).

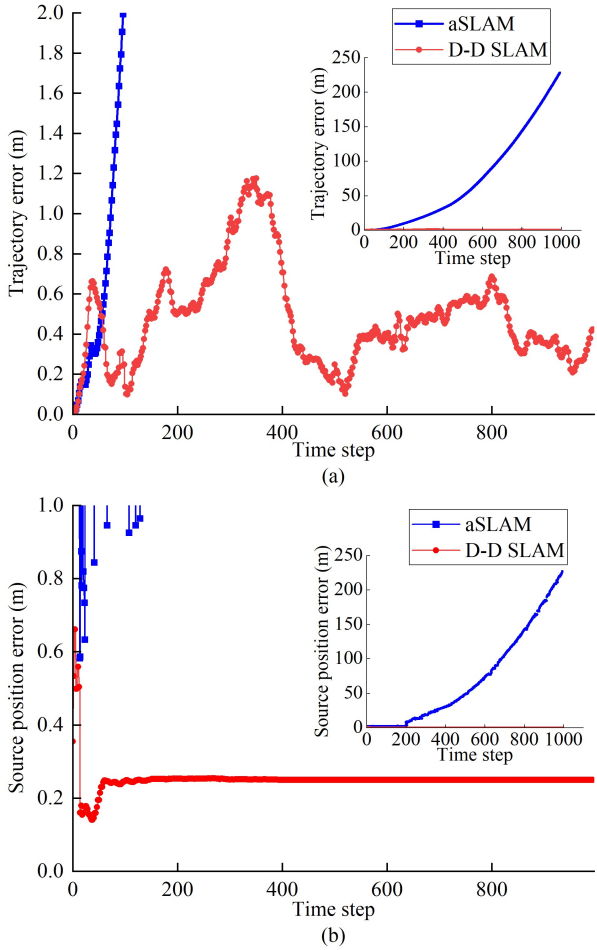

**Fig. 8.** Dataset result for (a) the trajectory error and (b) the source position estimation error over time steps.

Compared with aSLAM, the proposed D-D SLAM has an acceptable performance in the trajectory and source position error, proving its validity for real world indoor scenes. The results in Fig. 8 show the potential of the proposed D-D SLAM to work with an IMU in closed environments.

Fig. 9 shows the critical distance estimation on the DREGON dataset. Similar to that in the simulation, the estimated critical distance converges rapidly to a stable value near 6.8 m.

*C. Analysis*

In this section, the reasons for the trajectory errors are analyzed. The trajectory errors are affected by the number of particles, the max number of GM components, the DoA estimates, the IMU measurements, and the robot's distance from the source.

The positional errors in different particle numbers are observed (as triangle and circle marks in Fig.10(a)). It is also important to determine the influence of the particle number on the critical distance estimation. The experiment in section V part B is repeated for 5, 8, 10, 15 and 20 particles to investigate if a further improvement of the positional accuracy and the critical distance estimation can be achieved.

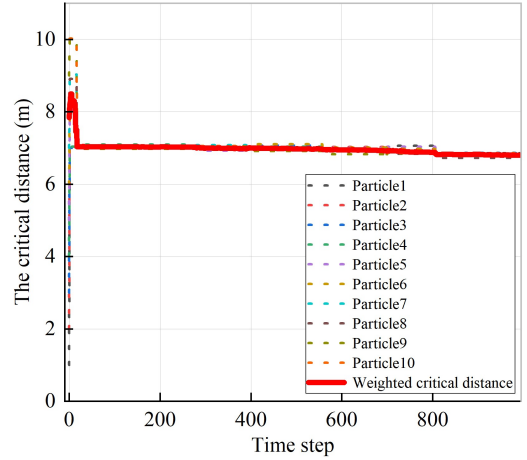

**Fig. 9.** The state of different critical distance particles and the weighted critical distance with real data.

When the number of particles increases from 5 to 10, the errors of source position and robot trajectory estimation decline quickly because the brute force strategy consists of increasing the number of particles and proves to be effective [32]. When the number of particles increases above 10, the errors of the source position and the robot trajectory estimation stabilize at values of 0.25 m and 0.48 m, respectively. The reason for this is that the contribution of increasing the particle number is negligible after a few iterations [33]. The critical distance estimation also quickly converges to a stable value for the same reason.

The experiment in section V part B is also repeated for different max numbers of GM components (as triangle and circle marks in Fig.10(b)). With an increase in the maximum of GM components, the positional error and the critical distance estimates quickly converge to a stable value.

As mentioned before, the SRP-PHAT with limited elevation search boundaries has high-accuracy DoA results. Even if the direction of the drone noise is in the range of the source direction, the Wiener Filter with the noise sample can be applied for noise reduction [15]. Incidentally, the drone noise can only affect the ITD and the coherence of binaural signals in the calculation process of DRR. As the ITD is calculated with accurate DoA estimates, the ITD is robust to drone noise. As the drone noise is diffuse and incoherent between different channels, it can be suppressed when calculating the coherence of binaural signals.

As the covariance of the DoA estimates and the IMU measurements are considered to be time invariant while the trajectory errors have obvious variation with time, the time-varying robot's distance from the source is considered. In Fig. 11, the true robot-source distance is represented with the blue dashed line, and the trajectory error of D-D SLAM is drawn with the red solid line. What can be Fig. 11 clearly shows is that the trajectory error increases with the decrease in the true robot-source distance. Roughly speaking, the trajectory error reaches the maximum when the true robot-source distance approaches the minimum. Therefore, it can be assumed that the trajectory error is associated with the robot-source distance.

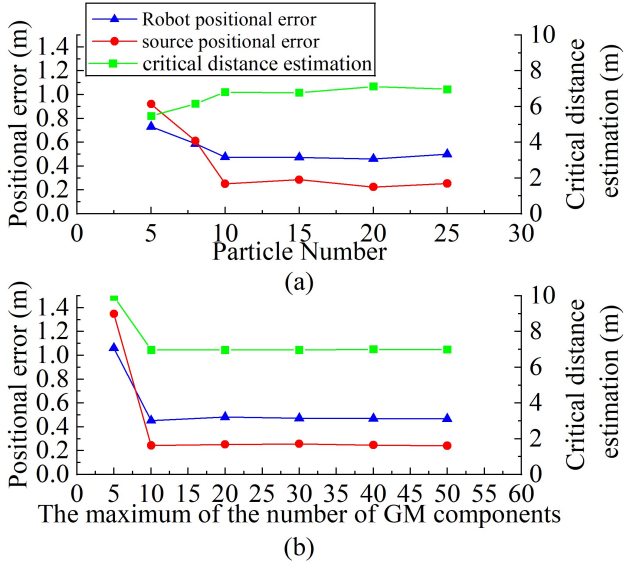

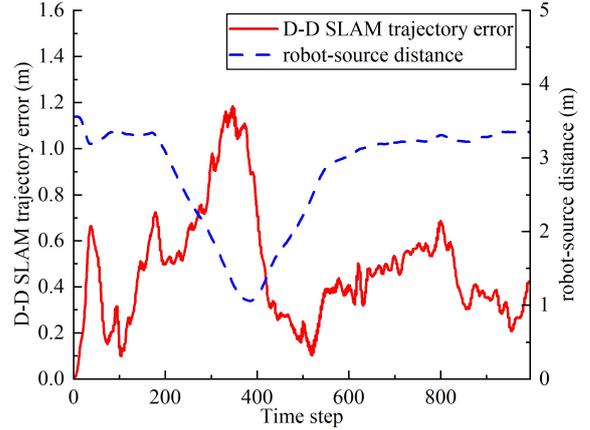

**Fig. 11.** The robot-source distance and the trajectory error of D-D SLAM.

**Fig.10.** The positional error and the critical distance estimation with different numbers of particles and the max number of GM components.

Using (16), the estimation of the robot-source distance depends on the critical distance. The critical distance should be fixed when the reverberation field is stable. However, the reverberation field may vary if there are moving objects. The closer the robot is to the source, the larger its influence on the reverberation field. The reason is that when they are close, the multiple diffractions and the multiple reflections they cause will become stronger, and the influence on the reverberation field is also heavier, and vice versa.

As shown in Fig. 9, when the robot is still far from the source, the critical distance estimation has already converged to a stable value, leading to particle dilution of the critical distance. After that, the variation in the reverberation field caused by the change in distance between the robot and the source can no longer be considered with the diluted particle filter. As a result, the outdated estimation of the critical distance will affect the estimation of the robot-source distance, according to (16). The deviation of the robot-source distance finally causes an increase in the trajectory estimation error.

## VII. CONCLUSION

Focusing on the solution of indoor SLAM with acoustic data and an IMU, D-D SLAM is proposed. With the keyframe method, D-D SLAM performs well in mapping regardless of whether the robot moves or remains stationary. The use of the DRR for the estimation of the source's distance from the robot as a new constraint in D-D SLAM effectively eliminates the nonlinear noise of the IMU measurements. As the key factor to calculate the source's distance from the DRR estimates, the critical distance is unknown initially, so a particle filter is applied to estimate the critical distance online, and the estimation of the critical distance converges to a stable value.

According to the results, D-D SLAM has good performance in both simulations and experiments. For the first time, an acoustic SLAM algorithm is validated with a real indoor dataset containing only acoustic data and IMU measurements. Different from previous work, D-D SLAM is designed for three-dimensional motion and rotation. On the real indoor dataset, the proposed D-D SLAM can locate the robot with an average accuracy of 0.48 m and build a source map with an average accuracy of 0.25 m. Even though the initial source position is random, the error of the source position converges to less than 0.25 m within 2.8 s.

These results demonstrate the effectiveness of the proposed D-D SLAM in real world indoor scenes. In the future, D-D SLAM may contribute to robot localization and map building when conventional optical sensors are not suitable for special environments, for example, foggy rooms where light and lasers cannot penetrate in search or rescue missions after disasters.

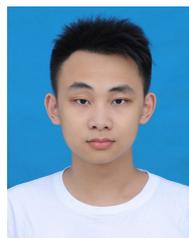

**Wenhao Qiu** received the B.E. degree in HNU College of Mechanical and Vehicle Engineering from Hunan University in 2019. He is now a Ph.D. candidate at the Hunan University with the bachelors degree lead to Ph.D. Programs. His current research interests include the robot audition and the acoustic SLAM.

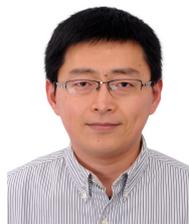

**Gang Wang** received the B.S. degree and the Ph.D. degree in Mechanical Engineering, College of Mechatronic Engineering and Automation, National University of Defense Technology, China in 1998 and 2006 respectively. He is a Professor in the State Key Laboratory of Advanced Design and Manufacturing for Vehicle Body, Hunan University, China now. Professor Wang has published over 60 papers. He received awards, including National Excellent Doctoral Dissertation Award (2008) and Natural Science Award of Hunan Province, China (2011). His current research interests include the acoustic signal processing for robot sensing systems.

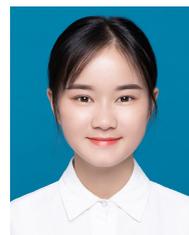

**Wenjing Zhang** received a B.E. degree in Mechatronics Engineering from Hebei University of Technology, China in 2020. She is now studying for a master's degree in Mechanical and Vehicle Engineering at Hunan University, China. Her current interests are acoustic SLAM on mobile robots.